\documentclass[letterpaper]{article} 
\usepackage{aaai2026}  
\usepackage{times}  
\usepackage{helvet}  
\usepackage{courier}  
\usepackage[hyphens]{url}  
\usepackage{graphicx} 
\usepackage{natbib}  
\usepackage{caption} 
\usepackage{algorithm}
\usepackage{algorithmic}

\usepackage{newfloat}
\usepackage{listings}
\DeclareCaptionStyle{ruled}{labelfont=normalfont,labelsep=colon,strut=off} 
\floatstyle{ruled}
\newfloat{listing}{tb}{lst}{}
\floatname{listing}{Listing}
\newcommand*{\data}{\ensuremath{\boldsymbol{x}}}

\usepackage{amsmath}
\usepackage{amsfonts}
\usepackage{multirow}
\usepackage{subcaption}
\usepackage{booktabs}

\title{Graph-Conditional Flow Matching for Relational Data Generation}
\author{
    Davide Scassola\textsuperscript{\rm 1,2},
    Sebastiano Saccani\textsuperscript{\rm 2},
    Luca Bortolussi\textsuperscript{\rm 1}
}
\affiliations{
    \textsuperscript{\rm 1}AI\textsc{lab}, University of Trieste, Trieste, Italy\\
    \textsuperscript{\rm 2}Aindo, AREA Science Park, Trieste, Italy\\
    davide.scassola@phd.units.it, sebastiano@aindo.com, lbortolussi@units.it
}

\usepackage{bibentry}

\begin{document}

\maketitle

\begin{abstract}
Data synthesis is gaining momentum as a privacy-enhancing technology.
While single-table tabular data generation has seen considerable progress, current methods for multi-table data often lack the flexibility and expressiveness needed to capture complex relational structures.
In particular, they struggle with long-range dependencies and complex foreign-key relationships, such as tables with multiple parent tables or multiple types of links between the same pair of tables.
We propose a generative model for relational data that generates the content of a relational dataset given the graph formed by the foreign-key relationships.
We do this by learning a deep generative model of the content of the whole relational database by flow matching, where the neural network trained to denoise records leverages a graph neural network to obtain information from connected records.
Our method is flexible, as it can support relational datasets with complex structures, and expressive, as the generation of each record can be influenced by any other record within the same connected component.
We evaluate our method on several benchmark datasets and show that it achieves state-of-the-art performance in terms of synthetic data fidelity.
\end{abstract}


 \begin{links}
     \link{Code}{https://github.com/DavideScassola/graph-conditional-flow-matching}
 \end{links}

\section{Introduction}

Data has become a fundamental resource in the modern world, playing an essential role in business, research and daily life. However, privacy concerns often restrict its distribution.
Since most data is stored in relational tables, synthetic data generation is emerging as a solution for sharing useful insights without exposing sensitive information. This approach can ensure compliance with privacy regulations such as the European Union’s General Data Protection Regulation (GDPR).

Tabular data synthesis \cite{ctgan, shi2025tabdiff} has been subject to research for several years. Early methods were based on Bayesian networks \cite{zhang2017privbayes}, factor graphs \cite{mckenna2019graphical} and autoregressive models \cite{nowok2016synthpop}. Most recent methods leverage the breakthroughs of deep-learning based generative models, from early latent variables models as VAEs \cite{kingma2013auto} and GANs \cite{goodfellow2014generative}, to transformer-based autoregressive models \cite{vaswani2017attention} and diffusion models \cite{ho2020denoising}.
Despite the advancements in single table synthesis, generating multiple tables characterized by foreign-key constraints is a considerably more difficult task.

Relational datasets can be represented as large graphs, where nodes correspond to records and edges denote foreign-key relationships. This introduces a dual challenge: (1) modeling potentially complex graph structures such as tables with multiple parents, or multiple types of relationships between two tables and (2) modeling statistical dependencies between records linked directly or indirectly through foreign keys.

Relational data generation \cite{patki2016synthetic, gueye2022row, solatorio2023realtabformergeneratingrealisticrelational, pang2024clavaddpm} is a less mature field, with few existing methods capable of properly handling complex database structures. In \citet{xu2022synthetic}, they separately model the graph structure and then propose a method for generating the content of the relational database table-by-table.
The generation of each record is conditioned on graph-derived node statistics and aggregated information from connected records.
In concurrent work, \citet{hudovernik2024relational} follow a similar approach, generating the content of the tables using a latent diffusion model conditioned on node embeddings produced by a separate model.

In this work, we propose a method for generating the content of a relational dataset given the graph describing its structure. Inspired by recent advancements in image generation,
we employ flow matching to train a flow-based generative model of the content of the entire relational dataset. In order to enable information propagation across connected records, the architecture of the learned denoiser includes a graph neural network (GNN) \cite{scarselli2008graph, gnn2}. 
This approach aims at maximizing expressiveness in modeling correlation between different records of the database, as information can be passed arbitrarily within a connected component through a GNN. Moreover, our framework is flexible as the conditioning graph can be complex, and scalable as we can generate large datasets.

Using SyntheRela \cite{hudovernik2024benchmarking}, a recently developed benchmarking library, we prove the effectiveness of our method on several datasets, comparing it with several open-source approaches. Experimental results show our method achieves state-of-the-art performance in terms of fidelity of the generated data.

\section{Background}
\label{background}

\subsection{Continuous Normalizing Flows}

Flow matching \cite{lipman2022flow} is an emerging framework for training continuous normalizing flows (CNFs), a deep generative model that learns to transform random noise into target distributions through ordinary differential equations.

Given data $\data \in \mathbb{R}^d$ sampled from an unknown data distribution $q(\data)$ and a time-dependent vector field $v : [0,1] \times \mathbb{R}^d \rightarrow \mathbb{R}^d $, also called \textit{velocity}, a continuous normalizing flow $\varphi : [0,1] \times \mathbb{R}^d \rightarrow \mathbb{R}^d $ is a transformation defined by the following ODE:
\[
\frac{d}{d t} \varphi_t(\data)  =v_t(\varphi_t(\data)) \text{ with initial conditions } \varphi_0(\data)  =\data
\]
This transformation can be used to map a given tractable distribution $p_0(\data)$ (e.g., a Gaussian) into a more complex distribution $p_1(\data)$. The family of density functions $p_t(\data)$ for $t \in [0,1]$ is known as the \textit{probability density path}, and $v_t$ is said to generate the probability density path $p_t(\data)$.

Flow matching provides a framework for training a deep neural network to parameterize a velocity field $v_t(\data)$ such that the resulting ordinary differential equation transforms samples from a tractable noise distribution $p_0$ into samples approximating the target data distribution $q \approx p_1$.

\subsection{The Flow Matching Objective}

Since direct access to the vector field $v_t$ of a CNF generating the data is unavailable, we cannot directly match a parameterized velocity field $v_t^{\theta}$ against the ground truth $v_t$.
The main idea of flow matching is instead to define the underlying probability path as a mixture of conditional "per-example" probability paths, that can be defined in a tractable way.

Let us denote by $p_t(\data \mid \data_1)$ a conditional probability path such that $p_0(\data \mid \data_1) = p_0(\data)$ and $p_1(\data \mid \data_1)$ is a distribution  concentrated around $\data = \data_1$, as $p_1(\data \mid \data_1) = \mathcal{N}(\data \mid \data_1, \sigma^2I)$ with $\sigma$ small.
We define the conditional velocity $u_t(\data \mid \data_1)$ as the velocity generating the conditional probability path $p_1(\data \mid \data_1)$.
It is then possible to prove that the marginal velocity, defined as

\begin{equation*}
  \begin{aligned}
    u_t(\data) & = \int u_t (\data \mid \data_1 ) p_t(\data_1 \mid \data) d \data_1\\
      & = \int u_t (\data \mid \data_1 ) \frac{p_t (\data \mid \data_1 ) q (\data_1 )}{p_t(\data)} d \data_1
  \end{aligned}
\end{equation*}

generates the marginal probability path
\[
p_t(\data)= \int p_t(\data \mid \data_1) q(\data_1) d\data_1
\]
that, following the definition of conditional probability path, at $t=1$ closely matches the target distribution $q(\data)$.
Moreover, it can be shown that a valid loss for learning the marginal velocity is the following:
\[
\mathcal{L}_{\mathrm{FM}}(\theta)=\mathbb{E}_{t \sim \mathcal{U}[0,1], \data_1 \sim q (\data_1 ), \data \sim p_t (\data \mid \data_1 )}\|v_t^\theta(\data)-u_t (\data | \data_1) \|^2
\]
known as the conditional flow matching objective, where a neural network learns to match the marginal velocity by matching conditional velocities.

\subsection{Optimal Transport Flows}
A common and effective choice of the conditional probability paths is the optimal transport (OT) displacement map between the two Gaussians $p_0(\data \mid \data_1) = \mathcal{N}(0,I)$ and $p_1(\data \mid \data_1)= \mathcal{N}(\data_1, \sigma_{min}I )$:
\[
p_t(\data \mid \data_1) = \mathcal{N}( t \data_1, I (1 - t + t \sigma_{min}))
\]
generated by the following conditional velocity
\begin{equation}
\label{eq:ot_velocity}
u_t(\data \mid \data_1)=\frac{\data_1-(1-\sigma_{\min }) \data}{1-(1-\sigma_{\min }) t}
\end{equation}

Alternatively, one can write $\data_t \sim  p_t(\data \mid \data_1)$  as $\data_t = t \data_1 + (1-t +t\sigma_{min}) \data_0 $ and $u_t(\data \mid \data_1) = \data_1 - \data_0 (1-\sigma_{min})$ with $\data_0 \sim \mathcal{N}(0,I)$.

\subsection{Variational Parametrization}
In \citet{eijkelboom2024variational}, they show an alternative to directly matching the conditional velocity. They propose the following parametrization of the learned marginal velocity

\begin{equation*}
  \begin{aligned}
    v_t^\theta(\data) & := \int u_t(\data \mid \data_1) q_t^\theta(\data_1 \mid \data) \mathrm{d} \data_1\\
      & = \mathbb{E}_{\data_1 \sim q_t^\theta(\data_1 \mid \data)} u_t(\data \mid \data_1)
  \end{aligned}
\end{equation*}
Then the marginal velocity can be learned by matching a variational distribution $q_t^\theta(\data_1 \mid \data)$ to the ground truth $p_t(\data_1 \mid \data)$ by minimizing $D_{\text{KL}}(p_t(\data_1 \mid \data) \| q_t^\theta(\data_1 \mid \data))$. This is equivalent to maximum likelihood training:
\[
\mathcal{L}_{\mathrm{VFM}}(\theta)=-\mathbb{E}_{t \sim \mathcal{U}[0,1], \data_1 \sim q (\data_1 ), \data \sim p_t (\data \mid \data_1 )}\log q_t^\theta(\data_1 \mid \data)
\]
If the conditional flow is linear in $\data_1$ (as in the case of the OT map), then matching the expected value of $p_t(\data_1 \mid \data)$ is enough to learn the marginal velocity $u_t(\data \mid \data_1)$. This implies that the variational approximation can be a fully factorized distribution, without loss of generality. In this work, we refer to the learned neural network as the \textit{denoiser}.

According to \citet{eijkelboom2024variational}, this parametrization offers advantages when dealing with one-hot-encoded categorical variables. First, the generative paths become more realistic due to the inductive bias, which improves convergence by avoiding misaligned paths. Second, using cross-entropy loss instead of squared error enhances gradient behavior during training, thereby speeding up convergence.

\begin{figure*}
    \centering
    \includegraphics[width=1\linewidth]{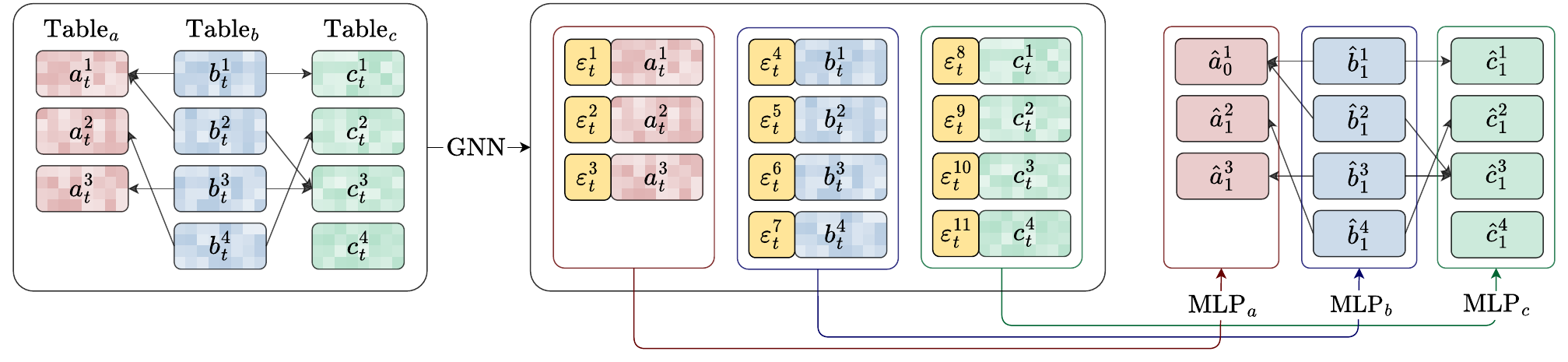}
    \caption{Overview of the architecture of the denoiser for relational data. A relational dataset composed of multiple tables can be seen as a graph, where records are the nodes and foreign keys are the edges. The denoiser takes as input a relational dataset where noise was added to each record with noise level $t$. Firstly, a graph neural network (GNN) processes the entire graph and computes node embeddings $\varepsilon^i_t$ encoding context information for each record. Each record and its corresponding embedding are then processed independently by table-specific multi-layer perceptrons (MLPs), which predict the original clean records ($t=1$).
    }
    \label{fig:cartoon}
\end{figure*}

\section{Method}

\subsection{Relational Data Generation}
Relational databases are composed of multiple tables, where each table is a collection of records with a common structure. Tables may include one or more columns containing foreign keys, allowing records to refer to records in other tables.
Consequently, a relational database can be represented as a graph, where nodes correspond to records and edges are defined by foreign-key relationships. In particular, the resulting graph is heterogeneous, since the nodes belong to different tables and have different types of features (i.e., the fields of a record).
By convention, if a table $A$ contains foreign keys referring to records in table $B$, $A$ is called the \textit{child} table and $B$ the \textit{parent} table.

Let $\data^i$ denote the features of node $i$, and by $g^i$ the set of nodes to which $i$ is connected to. Then a relational dataset $\mathbb{D}$ is identified by the pair $(\mathbb{X}, \mathbb{G})$, where $\mathbb{X} := \{\data^i\}_{i=1}^N$ and $\mathbb{G} := \{g^i\}_{i=1}^N$. We often refer to $\mathbb{X}$ as the \textit{content} of the relational database, and we refer to $\mathbb{G}$ as the \textit{foreign-key graph} or \textit{topology} of the relational database. Moreover, as nodes are grouped into $K$ tables $\mathbb{T}_k$ sharing the same features structure, we can also write $\mathbb{X} = \{\mathbb{T}_k\}_{k=1}^K$.

Several approaches have been explored for relational data generation that differ in the order in which tables and topology are generated. An approach that has been recently proven effective \cite{xu2022synthetic} is to first generate the topology and then generate the tables one by one, conditioning on the topology and on previously generated tables: 
\[
    p(\mathbb{X}, \mathbb{G}) = p(\mathbb{G}) \prod_{k=1}^K p(\mathbb{T}_k \mid \mathbb{T}_{1:k-1}, \mathbb{G})
\]
Our approach is similar in the sense that we first generate the topology, but we generate the features contained in the tables all at once:
\[
p(\mathbb{X}, \mathbb{G}) = p(\mathbb{G})p(\mathbb{X} \mid \mathbb{G})
\]

In this work we focus on the problem of conditional generation of the features $\mathbb{X}$ given the topology $\mathbb{G}$ of the relational database $p(\mathbb{X} \mid \mathbb{G})$.
In this way, the method for generating the foreign-key graph is independent from the method generating the features. This modular approach has advantages, as methods for generating large graphs are often quite various and different from deep generative models.
Moreover, the complexity of our conditional generation method scales linearly with the size of the graph, while most methods for graph generation scale quadratically \cite{zhu2022survey}. Finally, a conditional generation method is useful when one is interested in anonymizing the content but not the structure of a relational dataset, or when the topology is directly given by the user.

\subsection{Foreign-key Graph Generation}
Since our work focuses on conditional generation of relational data content given a fixed topology, we adopt a simplified sampling approach for the foreign-key graph $\mathbb{G}$ (treating topology generation as orthogonal to our core contribution).
For datasets where $\mathbb{G}$ has a large connected component, containing most or all nodes, we just keep the original topology $\mathbb{G}$. Otherwise, we build a new topology by sampling with replacement the connected components of $\mathbb{G}$ \cite{hudovernik2024relational}.
This method could be replaced by a dedicated graph sampler as in \citet{xu2022synthetic},  which we consider an interesting direction for future work.
We remark that the subject of this work is the conditional generation of the content given the foreign-key graph. An advantage of this approach is its modularity, as this method can be combined with any pure graph generation approach.
Moreover, resampling the connected components  has concrete applications, for instance, when all the observed topologies are well-represented in the training data. 

\subsection{Generative Modeling from Single-Sample Data}

We propose to learn a conditional generative model $p(\mathbb{X} \mid \mathbb{G})$ for the whole set of features $\mathbb{X}$, since in principle, it cannot always be decomposed into several independently and identically distributed (i.i.d.) samples, that in this case would correspond to the connected components of the graph.
For example, in the movielens dataset \cite{movielens}, almost all records belong to the same connected component. Single-sample scenarios are common in large graph generation problems (e.g., social networks). Time series are another example where identifying i.i.d. samples is problematic.
Nevertheless, successful modeling when disposing of only one sample remains feasible when the single sample is composed of weakly interacting components. This is the case of relational data, where records are usually not strongly dependent on all other records belonging to the same connected component. Moreover, the graph is sparse since the number of foreign keys is proportional to the number of records.
Our method exploits structural regularities to enable effective learning from what is essentially a single sample $\mathbb{D}=(\mathbb{X}, \mathbb{G})$. 

In order to avoid the trivial solution where the model simply learns to reproduce the only available training sample $\mathbb{X}$, we have to carefully handle overfitting. This ensures the generative model generalizes and learns meaningful regularities in the data rather than merely learning to copy the specific instance.

The main motivation for this approach was to develop a maximally expressive generative model for tabular data, addressing the limitations of existing methods.
In practice, we achieve this by learning a flow using a modular architecture for the denoiser. This is composed of one denoiser for each table and a GNN. The GNN computes node embeddings for each record, encoding context information. Then, node embeddings are passed to the table-specific denoisers. In this way, the denoising process of each record is made dependent on the other connected records.

\subsubsection{Graph-Conditional Flow Matching}

In order to model $p(\mathbb{X} \mid \mathbb{G})$ using flow matching, we have to define the conditional flow $p_t(\mathbb{X} \mid \mathbb{X}_1, \mathbb{G})$.
We use the optimal transport conditional flow described in Section \ref{background} as conditional flow $p_t(\data^i \mid \data_1^i)$, independent for each node $\data^i \in \mathbb{X}$ and for each component of a node (for each field of each record).
We refer to the relative conditional velocity as $u_t(\mathbb{X} \mid \mathbb{X}_1)$. This also holds for categorical components of features $\data^i$, which are encoded in continuous space using one-hot encodings \cite{eijkelboom2024variational}. Notice that this conditional flow does not depend on the topology $\mathbb{G}$, then we can refer to it as $p_t(\mathbb{X} \mid \mathbb{X}_1)$.
The objective is to learn the marginal velocity of the whole relational dataset $v_t(\mathbb{X} \mid \mathbb{G})$. 
We learn this using the variational parametrization discussed above:
\[
p_\theta(\mathbb{X}_1 \mid \mathbb{X}_t, \mathbb{G}) \approx q(\mathbb{X}_1 \mid \mathbb{X}_t, \mathbb{G}) 
\]
The training loss is then the following:
\[
    \mathcal{L}(\theta) = \mathbb{E}_{t \sim \mathcal{U}[0,1], \mathbb{X}_t \sim p_t (\mathbb{X}_t \mid \mathbb{X}_1 )} \left[ - \log p_\theta(\mathbb{X}_1 \mid \mathbb{X}_t, \mathbb{G}) \right]
\]
Where $(\mathbb{X}_1, \mathbb{G})$ is the original relational dataset. Since the relational dataset $\mathbb{D} =(\mathbb{X}, \mathbb{G})$ is both the dataset and the only sample, using this loss corresponds to doing full-batch training.
However, it is also possible to write the loss as an expectation over the different connected components of $\mathbb{D}$ when possible.
Finally, the velocity used at generation time is the following:
\begin{equation}
\label{eq:rd_velocity}
    v_t^\theta(\mathbb{X}, \mathbb{G}) = \mathbb{E}_{\mathbb{X}_1 \sim p_\theta(\mathbb{X}_1 \mid \mathbb{X}, \mathbb{G})} \left[ u_t(\mathbb{X} \mid \mathbb{X}_1, \mathbb{G}) \right]
\end{equation}

\subsubsection{Variational Parametrization}
Let $x^{k,d,j}$ be the $j_{th}$ value of column $d$ of table $k$, then we can write
$\mathbb{X} = \left\{ x^{k,d, j} \;\middle|\; 
k = 1, \dots, K;\; 
d = 1, \dots, D_k;\; 
j = 1, \dots, N_k
\right\}$,
where $D_k$ and $N_k$ are respectively the number of columns and records in table $k$.
We use a fully factorized distribution as variational approximation. This means that the distribution factorizes into an independent distribution for each component $x^{k,d,j}$.
Then we can write the variational approximation as
\[
    p_\theta(\mathbb{X}_1 \mid \mathbb{X}_t, \mathbb{G}) = \prod_{k=1}^K \prod_{ d=1}^{D_k} \prod_{ j=1}^{N_k} p^{k,d}_\theta (x_1^{k,d, j} \mid \mathbb{X}_t, \mathbb{G})
\]
where $p^{k,d}$ represents the variational factor corresponding to column $d$ of table $k$.

Depending on the nature of variable $x^{k,d, j}$, continuous or categorical, we parametrize a different distribution $p^{k,d}$. In both cases, the distribution is parametrized by a trainable neural network denoiser composed of two modules:
\begin{enumerate}
\item A graph neural network $\eta_{\theta_1}$ that computes node embeddings $\varepsilon^i_t = \eta_{\theta_1}(\mathbb{G}, \mathbb{X}_t)^i$ for each node $\data^i$, encoding context information.
\item Feedforward neural networks $f^{k,d}_{\theta_2}(x^{k,d, j}_t, t, \varepsilon^i_t)$ using noisy records $x^{k,d, j}_t$, time (noise level) $t$, and node embeddings $\varepsilon^i_t$ to parametrize the distribution $p^{k,d}$.
\end{enumerate}

\paragraph{Categorical Variables.}
When component $x^{k,d,j}$ is categorical, we use a categorical distribution:
\begin{multline*}
    p^{k,d}_\theta (x^{k,d, j}_1 \mid \mathbb{X}_t, \mathbb{G}) =\\ \mathrm{Categorical} \left(x^{k,d, j}_1 \mid  \mathbf{p}=f^{k,d}_{\theta_2}(x^{k,d, j}_t, t, \varepsilon^i_t) \right)
\end{multline*}
For categorical variables, the last layer is a softmax function and training corresponds to training a neural network to classify $x^{k,d, j}_1$ using cross-entropy loss.

\paragraph{Continuous Variables.}
When component $x^{k,d,j}$ is a real number, we use the same architecture to parametrize the mean of a normal distribution with unit variance:
\begin{multline*}    
    p_\theta(x^{k,d,j}_1 \mid \mathbb{X}_t, \mathbb{G}) =\\ \mathcal{N}(x^{k,d,j}_1 \mid \mu=f^{k,d}_{\theta_2}(x^{k,d,j}_t, t, \varepsilon^i_t), \sigma=1)
\end{multline*}
In this case training corresponds to training a neural network regressor to predict $x^{k,d,j}_1$ using the squared error loss. We use a fixed unit variance since learning the mean is sufficient to correctly parametrize the velocity field \cite{eijkelboom2024variational}.

\paragraph{Velocity Computation.} The velocity for each component $x^{k,d,j}$ at time $t$ then follows from Equation \ref{eq:ot_velocity} and \ref{eq:rd_velocity}:
\begin{multline*}
    v^\theta_t(x^{k,d,j} \mid \mathbb{X}_t, \mathbb{G}) = \\
    \frac{\mathbb{E}_{x^{k,d,j}_1 \sim p_\theta^{k,d}(x_1^{k,d, j} \mid \mathbb{X}_t, \mathbb{G})} [x_1^{k,d, j}] - (1-\sigma_{\min}) x^{k,d,j}}{1 - (1-\sigma_{\min}) t}
\end{multline*}
Since we directly parametrize the mean of the distribution in both the categorical and the continuous case, we can just plug the output of the denoiser:
\begin{multline}
\label{eq:x_velocity}
    v^\theta_t(x^{k,d,j} \mid \mathbb{X}_t, \mathbb{G}) = \\
    \frac{ f^{k,d}_{\theta_2}(x^{k,d, j}_t, t, \varepsilon^i_t)-(1-\sigma_{\min }) x^{k,d,j}}{1-(1-\sigma_{\min }) t}
\end{multline}
Recall that $x^{k,d,j}$ can be either a continuous variable or a one-hot-encoded categorical variable. In this latter case, both $x^{k,d,j}$ and $v^\theta_t(x^{k,d,j} \mid \mathbb{X}_t, \mathbb{G})$ are vectors.


\paragraph{Architecture.} Notice that the neural network $f_{\theta_2}^{k,d}$ is specific to the column $d$ of table $k$. In particular, we use a different multi-layer perceptron for each table $k$, with a prediction head (the last linear layer) for every component $d$. The inputs of $f_{\theta_2}^{k,d}$ are concatenated and flattened.

Instead, the neural network $\eta_{\theta_1}$ computing node embeddings $\varepsilon^i_t =\eta_{\theta_1}(\mathbb{G}, \mathbb{X}_t)^i$, refers to a single GNN supporting heterogeneous graph data, i.e., graphs with multiple types of nodes, and as a consequence, different types of edges. We experimented with architectures based on GIN \cite{xu2018powerful} and GATv2 \cite{brody2021attentive}. In order to build a GNN compatible with heterogeneous graphs, we take an existing GNN model and use a dedicated GNN layer for each edge type, as showed in the documentation of the torch-geometric library \cite{pytorch_geometric_hetero}. The messages produced by each edge-specific layer need to be of the same dimension, so that they can be summed together to obtain a node embedding. A detailed description of the employed GNN architectures is provided in Appendix \ref{app:details}.
Figure \ref{fig:cartoon} shows an overview of the denoiser's architecture. 

\subsubsection{Computational Complexity}
Similarly to generative models for single-table data, the complexity of both training and generation of this method scales linearly with the number of records in the relational dataset.
First of all, the computational complexity of each layer of the GNN scales linearly with the number of edges. In a relational dataset the total number of foreign keys is proportional to the number of records, since each record (node) has a fixed number of foreign keys (edges). Consequently, the GNN's computational complexity is linear with the number of records.
Second, the multi-layer perceptrons are applied independently to each record, meaning their computational complexity also scales linearly and can be efficiently parallelized.

\begin{algorithm}[tb]
\caption{Graph-Conditional Generation}
\label{alg:sampling}
\textbf{Input}: $\mathbb{G} \sim p(\mathbb{G})$\\
\textbf{Parameter}: Number of steps $T$
\begin{algorithmic}[1] 
\STATE Sample $\mathbb{X}_0 \sim \mathcal{N}(0,I)$
\FOR{$t = 0$ \textbf{to} $1$ \textbf{step} $\frac{1}{T}$}
  \STATE $x_{t+\frac{1}{T}}^{k,d,j} = x_t^{k,d,j} + \frac{1}{T} v^\theta_t(x_t^{k,d,j} \mid \mathbb{X}_t, \mathbb{G})$ \hfill Eq.~(\ref{eq:x_velocity})
\ENDFOR

\STATE \textbf{return} $(\mathbb{X}_1, \mathbb{G})$
\end{algorithmic}
\end{algorithm}

\subsubsection{Implementation Details}

\paragraph{Data Preprocessing.}
Following \citet{kotelnikov2023tabddpm}, during the preprocessing phase we transform continuous features using quantile transformation, so that all marginals of continuous features will be normally distributed.
In order to handle missing data in numerical columns, we augment the tables including an auxiliary column containing binary variables indicating if the data is missing, and we fill missing values with the mean. This allows to preserve the information about missing data, and replicate missing data patterns in the generated samples. For categorical columns with missing data, we simply include a new category "NaN".
Tables that contain only foreign-key columns (and no features) are considered only when computing node embeddings.
Time embedding is implemented according to \citet{dhariwal2021diffusion}.

\paragraph{Training.}
In order to avoid overfitting, we randomly split nodes into a training and validation set, computing the loss only on training nodes. The experimental results report performance relative to models achieving the best validation loss during training.

We performed full-batch training, as the entire dataset and the denoiser's computations could be fitted into memory. This allowed us to denoise all record in parallel.
For scenarios involving significantly larger datasets, where the graph or the GNN computations exceed available memory, one could employ mini-batching techniques for GNNs, or strategies for scaling to out-of-memory graphs.

We observed relatively short training time: for the largest dataset, training took less than 20 minutes on a single GPU (see Appendix \ref{app:computational_resources} for more details), for the other datasets just a few minutes.

\paragraph{Generation.}
To solve the ODE generating the data, we used the Euler integration method with 100 steps. In our experiments the generation process was relatively fast: for the largest dataset we experimented with, the generation process took less than 10 seconds. We summarize in Algorithm \ref{alg:sampling} the graph-conditional generation algorithm.

\paragraph{Hyperparameters.}
In our experiments, we tuned hyperparameters to some extent depending on the dataset. These primarily include neural network parameters, such as the number of hidden units in feedforward layers. The size of the node embeddings is also a key hyperparameter. Tuning this value was important to balance expressiveness and overfitting, as overly large embeddings can lead the model to memorize structures. Across all experiments, we constrained the embedding size to values between 2 and 10. In Appendix \ref{app:details} we discuss how the validation loss changes as a function of this hyperparameter for two of the datasets.

\section{Experiments}

\subsection{Experimental Settings}
We evaluate our method using SyntheRela (Synthetic Relational Data Generation Benchmark)\cite{hudovernik2024benchmarking}, a recently developed benchmark library for relational database generation. This tool enables comparison of synthetic data fidelity (i.e., similarity to original data) across multiple open-source generation methods and various datasets.

\paragraph{Datasets.}
We experiment with six real-world relational datasets: AirBnB \cite{airbnb}, Walmart \cite{walmart}, Rossmann \cite{rossmann}, Biodegradability \cite{biodegradability}, CORA \cite{cora} and the IMDb MovieLens dataset \cite{harper2015movielens, movielens}, a commonly used dataset to study graph properties, containing users' ratings to different movies. The last three  dataset have tables with multiple parents. Some of these datasets were subsampled to enable comparison with other methods. More details are provided in Appendix \ref{app:datasets}.

\paragraph{Metrics.}
Our primary objective is generating high-fidelity synthetic relational data. To evaluate fidelity, we adopt a discriminator-based approach where an XGBoost classifier \cite{chen2016xgboost} (often the preferred classifier for tabular data) is trained to distinguish real from synthetic records. Lower discriminator accuracy indicates higher synthetic data quality, with an accuracy of $0.5$ implying indistinguishability.
While this is straightforward for single table datasets, where the input of the discriminator is a single row, this is not for relational data.
For relational data, we use the SyntheRela library's Discriminative Detection with Aggregation (DDA) metric, which extends single-table discrimination by enriching the content of the rows of parent tables with aggregate information of its "child" rows. In particular, they add to each row of a parent table the count of children, the mean of real-valued fields of children and the count of unique values of categorical value fields. 
We believe that discriminator-based metrics are a simple, concise and powerful way to evaluate the fidelity of synthetic data. We compare our method against those present in SyntheRela, that are the leading open-source approaches for relational data generation.

\subsection{Results}
We generate synthetic data for six of the datasets included in the SyntheRela library, and compare it with other relational data generation methods. In particular, we measure the accuracy of an XGBoost discriminator in the setting described above. Table \ref{tab:results} shows the average accuracy across different runs for each combination of dataset and method when possible. Where the dataset has multiple parent tables, the highest accuracy is reported.

Our method generally outperforms all baselines, often by a large margin. Moreover, it is applicable to all relational datasets considered, as it can handle complex schema structures, including tables with multiple parent tables, multiple foreign keys referencing the same table, and missing data. Missing results for some baselines are due to their limitations: ClavaDDPM cannot synthesize CORA and Biodegradability, as it only supports a single foreign-key relation between two tables, REaLTabF does not support tables with multiple parents and SDV fails to synthesize the IMDB dataset due to scalability issues \cite{hudovernik2024benchmarking}.
The performance metrics for other methods are taken from \citet{hudovernik2024relational}, where the SyntheRela library was also used for fidelity evaluation. Variability in the reported results is due to different initialization seeds used both for training and generation.

To assess the impact of the embeddings produced by the GNN, we evaluate the performance of our method when the embeddings are ablated. This corresponds to training separate single-table models for each table. The results were negatively affected, with the only exception being the CORA dataset. Nevertheless, we observed that ablating the GNN always led to a significant increase in validation loss, thus suggesting potential limitations of the discriminative metric in evaluating certain datasets. 

\begin{table*}
\centering
\begin{tabular}{lcccccc}
\toprule
 & \textbf{AirBnB} & \textbf{Biodegradability} & \textbf{CORA} & \textbf{IMDB} & \textbf{Rossmann} & \textbf{Walmart}\\
\midrule
Ours & $\mathbf{0.58 \pm 0.03}$ & $\mathbf{0.59 \pm 0.02}$ & $0.63 \pm 0.02$ & $\mathbf{0.59 \pm 0.03}$ & $\mathbf{0.51 \pm 0.01}$ & $\mathbf{0.73 \pm 0.01}$ \\
Ours (no GNN) & $0.70 \pm 0.005$ & $0.86 \pm 0.004$ & $0.62 \pm 0.004$ & $0.89 \pm 0.002$ & $0.75 \pm 0.01$ & $0.91 \pm 0.04$ \\
\citet{hudovernik2024relational} & $0.67 \pm 0.003$ & $0.83 \pm 0.01$ & $\mathbf{0.60 \pm 0.01}$ & $0.64 \pm 0.01$ & $0.77 \pm 0.01$ & $0.79 \pm 0.04$ \\
ClavaDDPM & $\approx 1$ & - & - & $0.83 \pm 0.004$ & $0.86 \pm 0.01$ & $0.74 \pm 0.05$ \\
RCTGAN  & $0.98 \pm 0.001$ & $0.88 \pm 0.01$ & $0.73 \pm 0.01$ & $0.95 \pm 0.002$ & $0.88 \pm 0.01$ & $0.96 \pm 0.02$ \\
REaLTabF. & $\approx 1$ & - & - & - & $0.92 \pm 0.01$ & $\approx 1$ \\
SDV & $\approx 1$ & $0.98 \pm 0.01$ & $\approx 1$ & - & $0.98 \pm 0.003$ & $0.90 \pm 0.03$ \\
\bottomrule
\end{tabular}
\caption{Average accuracy with standard deviation of an XGBoost multi-table discriminator using rows with aggregated statistics. For datasets with multiple parent tables, the highest accuracy was selected. The CORA dataset is the only one for which using GNN embeddings does not improve the evaluation metric. However, we noticed that the simple post-processing step consisting of removing duplicated records from a child table ($\approx 3\%$ of records), allowed us to obtain a performance of $\approx 0.50$. Moreover, we observed a lower validation loss when the GNN was used. Statistics are computed over three different runs.}
\label{tab:results}

\end{table*}

\paragraph{Privacy Evaluation.}
We evaluated potential privacy leaks in each table, where parent tables were enriched with aggregated information as previously described.
For each table, we computed the distance-to-closest-record (DCR) \cite{dcr2, dcr3} of each synthetic and real record comparing to a hold-out set of real records. We consider a synthetic table to exhibit privacy leakage if the percentage of its DCRs falling below the $\alpha$-percentile of the DCRs of real data is significantly greater than $\alpha$ \cite{quantile1, quantile3}. Intuitively, this indicates that synthetic records are close to real records more often than expected, suggesting potential privacy risk.
As shown in Table~\ref{tab:privacy_results}, when considering the $2\%$ percentile this percentage ($p_{\leq2\%}$) remains close to the expected value of $2\%$. The privacy score \cite{palacios2025contrastive}, a derived statistic that takes a value of zero (or slightly lower) when no privacy risk is detected and one when all synthetic records are deemed risky, is consistently close to zero, indicating negligible privacy risk in the  content of the synthetic relational data.

\begin{table*}

\centering

\begin{tabular}{l l r r c r}
\toprule
\textbf{Dataset} & \textbf{Table} & \textbf{\# Records} & \textbf{\# Features} & $\mathbf{p_{\leq2\%}}$ & \textbf{Privacy Score} \\
\midrule
\multirow{1}{*}{AirBnB} 
  & users    & 10,000  & 22 & $1.95\% \pm 0.08\%$ & $-0.0005 \pm 0.001$ \\

\cmidrule(lr){1-6}
\multirow{1}{*}{Biodegradability} 
  & molecule & 328     & 6  & $0.00\% \pm 0.00\%$ & $-0.02 \pm 0.000$ \\

\cmidrule(lr){1-6}
\multirow{2}{*}{IMDB MovieLens} 
  & movies   & 3,832   & 11 & $2.44\% \pm 0.04\%$ & $0.005 \pm 0.000$ \\
  & users    & 6,039   & 6  & $2.29\% \pm 0.21\%$ & $0.003 \pm 0.002$ \\

\cmidrule(lr){1-6}
\multirow{1}{*}{Rossmann} 
  & store    & 1,115   & 17 & $2.94\% \pm 0.26\%$ & $0.01 \pm 0.003$ \\

\cmidrule(lr){1-6}
\multirow{2}{*}{Walmart} 
  & depts    & 15,047  & 5  & $1.01\% \pm 0.04\%$ & $-0.01 \pm 0.000$ \\
  & features & 225     & 12 & $1.33\% \pm 1.93\%$ & $-0.007 \pm 0.020$ \\

\bottomrule
\end{tabular}
\caption{ Privacy results for tables having at least 100 records, at least 2 columns (after aggregation) and no more than 10\% of real DCRs equal to zero. Statistics are computed over three different runs.}
\label{tab:privacy_results}

\end{table*}

\section{Related Works}

\paragraph{Single Table Generation.} Early approaches for tabular data generation include Bayesian networks \cite{zhang2017privbayes}, autoregressive models \cite{nowok2016synthpop} and factor graphs \cite{mckenna2019graphical}, that usually required data to be discretized. Early deep latent‐variable models as VAEs and GANs were later extended to model heterogeneous tabular data \cite{ctgan, tgan}. More recently there have been works leveraging transformer-based deep autoregressive model \cite{castellon2023dp} and diffusion models \cite{kotelnikov2023tabddpm, zhang2023mixed, shi2025tabdiff}. Notably, in \citet{jolicoeur2024generating} they use flow matching where the denoiser is based on trees, which have often been shown to be state of the art for several tasks relating to tabular data \cite{grinsztajn2022tree, shwartz2022tabular, borisov2022deep}.

\paragraph{Relational Data Generation.} Synthesizing relational data introduces the additional challenge of preserving inter‑table dependencies. 
The Synthetic Data Vault (SDV) \cite{patki2016synthetic} pioneered multi‑table synthesis via the Hierarchical Modeling Algorithm (HMA), which employs Gaussian copulas and recursive conditional parameter aggregation to propagate child‐table statistics into parents.
GAN‑based relational extensions include \citet{gueye2022row}, conditioning child‑table synthesis on parent and grandparent row embeddings, and \citet{li2023irg}, where the generation of single rows is based on GANs, but tables are generated sequentially in an autoregressive way, following the foreign‑key topology.
Following a similar principle  \citet{solatorio2023realtabformergeneratingrealisticrelational} and \citet{gulati2023tabmt} leverage instead a transformer-based autoregressive model. 
These methods still cannot properly manage the generation of tables with multiple parents, since the number of children per parent's row depends only on one of the parents. In general, these sequential methods can only properly deal with tree-like topologies.
\citet{pang2024clavaddpm} introduce a guided diffusion approach using latent variables to capture long-range dependencies across tables. To handle tables with multiple parents, they generate a version of the child table for each parent and heuristically merge them by selecting similar rows.
In \citet{xu2022synthetic} the graph is first generated with a statistical method, and then the content of each table is sequentially generated by conditioning to the content of already generated tables and topological information.
A similar approach is proposed in concurrent and independent work by \citet{hudovernik2024relational}, sharing similarities with ours. Data generation is based on latent-diffusion, conditioned on a pre-generated graph by node embeddings encoding topological and neighborhood information, computed using a GNN.
Our work differs in three aspects: (1) we employ flow matching rather than latent diffusion; (2) our GNN is integrated into the denoiser, so it is trained end-to-end, whereas theirs uses embeddings precomputed independently from the generative models of the records; (3) we generate tables in parallel rather than sequentially.


\section{Limitations}
Like other generative models, our approach requires dataset-specific hyperparameter tuning, particularly the depth and width of the employed neural networks, to adapt to the training data size and avoid overfitting.
Although the architecture we used is relatively simple, we were able to achieve state-of-the-art performance. Therefore, we believe there is significant room for improvement through more sophisticated model design.
This work focuses on the method for training a flexible and powerful generative model for relational data, rather than on the specific architecture of the denoiser.

\section{Conclusion}


We proposed a novel approach for generating relational data, given the graph describing the foreign-key relationships of the datasets.
Our method uses flow matching to build a generative model of the whole content of a relational dataset, exploiting a GNN to increase the expressiveness of the denoiser, by letting information flow across connected records.
Our method achieves state-of-the-art performance in terms of synthetic data fidelity across several datasets, outperforming other open-source methods in the SyntheRela benchmark library.
Moreover, we did not observe any privacy leakage in the generated synthetic tables, even when parent records were enriched with aggregated statistics from their child tables.

\paragraph{Future Works.}
An interesting direction of development is the combination with generative models for large graphs, such as exponential random graph models \cite{robins2007introduction}. We think this approach is promising as the computational complexity of our method scales linearly with the size of the dataset, while deep generative models of graphs often scale quadratically \cite{zhu2022survey}. Moreover, foreign-key graphs are often simple enough to be modeled by less powerful but scalable statistical models.
Finally, as our method is based on flow matching, one can further exploit properties of diffusion-like models, such as guidance, inpainting, or the generation of variations of a given dataset. 


\section*{Acknowledgements}
This research was supported by Aindo, which has funded the PhD of the first author and provided the computational resources. Additionally, this study was carried out within the PNRR research activities of the consortium iNEST funded by the European Union Next-GenerationEU (PNRR, Missione 4 Componente 2, Investimento 1.5 – D.D. 1058 23/06/2022, ECS\_00000043).

\bibliography{aaai2026}

\newpage
\appendix

\appendix
\section{Computational Resources}
\label{app:computational_resources}

The models we trained are characterized by relatively fast training and sampling phases. On a single GPU, generation took at most a few seconds per model, and training took no more than 15 minutes. Table~\ref{tab:experiment_duration} reports the maximum runtime across repetitions for each dataset, including both training and generation.

\begin{table}[h]
\centering
\caption{Maximum runtime across repetitions for each dataset during experimentation.}
\label{tab:experiment_duration}
\begin{tabular}{l r}
\toprule
\textbf{Dataset Name} & \textbf{Running Time} \\
\midrule
AirBnB                & 10m 3s \\
Biodegradability      & 1m 6s  \\
CORA                  & 3m 10s \\
IMDB MovieLens        & 14m 25s \\
Rossmann              & 2m 57s \\
Walmart               & 1m 48s \\
\bottomrule
\end{tabular}
\end{table}

Computing the DDA metric using the \texttt{SyntheRela} library took less than one minute for each dataset. Considering three runs for every experiment, including training, generation, and metric evaluation, the total runtime was approximately two hours. The overall research project required more computation time due to the experiments involved in developing and empirically validating the method.

The following hardware was used to train the models and evaluate the results:
\begin{itemize}
    \item \textbf{Processor:} AMD Ryzen Threadripper 2950X (16-Core)
    \item \textbf{Memory:} 125 GiB
    \item \textbf{GPU:} NVIDIA RTX A5000
\end{itemize}

\section{Technical Details on Training, Architecture, and Hyperparameters}
\label{app:details}

We provide here a small overview of the technical details regarding neural network training and architecture, for more details, we refer the reader to the released code.

\subsection{Training Details}

As discussed earlier, training corresponds to maximum likelihood training, where the target is the original relational dataset, in the form of a graph, and the input is the relational dataset after noise is added according to the conditional probability path.
In each epoch, we performed $n$ loss evaluations and optimization steps. Every loss evaluation in an epoch is characterized by a different noise level $t$, in particular, we used $n$ equally spaced values in the interval $[0,1]$, with $n=100$ or $200$, depending on the experiment.

We use RAdam as the optimizer, together with an exponentially decaying learning rate scheduler, starting from approximately $10^{-3}$ and decaying to approximately $10^{-5}$.

Models are trained for 10 to 40 epochs, with early stopping based on validation performance.

\subsection{GNN Architectures}
\label{app:gnn_architectures}

We use graph neural networks (GNNs) to process relational data and compute node embeddings $\varepsilon^i$\footnote{In this section, we omit the time dependency $t$ to keep the notation uncluttered} for each node $i$ in the graph.

Assume the dataset consists of nodes $\data^i$ each one belonging to one of the $K$ tables. We refer the table $\data^i$ belongs to as $k_i$.
Each GNN layer computes a new representation $h^i$ for node $i$ as a function of its own features and those of its neighbors $N_i$:
\[
h^i = \text{GNNConv}(\data^i, N_i)
\]
Since the graphs are heterogeneous, with multiple node types and edge types, standard GNN layers must be extended to handle this structure. Following the design pattern in PyTorch Geometric for heterogeneous message passing \cite{pytorch_geometric_hetero}, we define the modified layer as:
\[
h^i = \sum_{k=1}^{K} \text{GNNConv}_{k \rightarrow k_i}(\data^i, N_i^{(k)})
\]
where $N_i^{(k)}$ is the set of neighbors of node $i$ that belong to node type $k$, and each $\text{GNNConv}_{k \rightarrow k_i}$ is an edge-type-specific instance of the base convolution layer. For brevity, we omit this heterogeneity adaptation in the main text but assume it in all GNN layers.

\paragraph{GATv2-based Architecture.}
Our primary GNN architecture is based on the GATv2 convolution layer \cite{brody2021attentive}. Each GNN block contains a GATv2 convolution followed by a residual linear transformation and ReLU activation:
\[
h_i = \text{ReLU}\left( \text{GATv2Conv}(\data^i, N_i) + \text{Linear}_{k_i}(\data^i) \right)
\]
\[
\varepsilon^i = \text{GATv2Conv}(h^i, N_i) + \text{Linear}(h^i)
\]
The hidden dimensionality is shared across all node types, while the input linear layers are type-specific. The only architecture-specific hyperparameter is the dimensionality of hidden layers $h^i$, which we set to 100 in our experiments.

\paragraph{GIN-based Architecture.}
We also experiment with a variant based on the Graph Isomorphism Network (GIN) \cite{xu2018powerful}. Each node is first projected into a shared latent space using a type-specific linear layer:
\[
\mathbf{z}^i = \text{Linear}_{k_i}(\data^i)
\]
We then apply multiple GINConv layers adapted to heterogeneous graphs as described above. The architecture-specific hyperparameters in this case are the size of embeddings $\mathbf{z}^i$, the number of GIN layers, and the width of the MLPs used in the GINConv modules. In our experiments, we used three GIN layers with an MLP width of 100. The size of the linear embedding was set to 20 or 50, depending on the experiment.

We employed the GATv2-based GNN for the AirBnB, Rossmann and Walmart datasets, while we used the GIN for the CORA, IMDB MovieLens, and Biodegradability datasets.

\subsection{Table-specific Denoisers}
Once node embeddings $\varepsilon^i_t$ are computed for each node $\data^i_t$, we use table-specific denoisers $f^{k}$ to parametrize the variational distributions. In our case, this corresponds to computing the expected value of the predictive distribution:
\[
\hat{\data}^i_1 = f^{k_i}(\data^i_t, t, \varepsilon^i_t)
\]

Each denoiser $f^k$ is implemented as a multi-layer perceptron (MLP), where the input is the concatenation of three components: the current noisy value $\data^i_t$, the time $t$ (embedded following \citet{dhariwal2021diffusion}), and the node embedding $\varepsilon^i_t$.

We apply Layer Normalization \cite{ba2016layer} after each hidden activation (SiLU \cite{elfwing2017silu}), except in the final layer. The output layer is linear and restores the original dimensionality of $\data^i_t$. For one-hot-encoded categorical features, a final softmax activation is applied.

For each table $k$, the hyperparameters are the number of layers and the width (i.e., number of hidden units) of each  layer in the MLP.
In our experiments we used 2 or 3 hidden layers, with the number of hidden units ranging from 10 to 1000 depending on the experiment.

\subsection{Tuning Node Embedding Size}
\begin{figure}[ht]
    \centering
    \begin{subfigure}[b]{0.45\textwidth}
        \includegraphics[width=\textwidth]{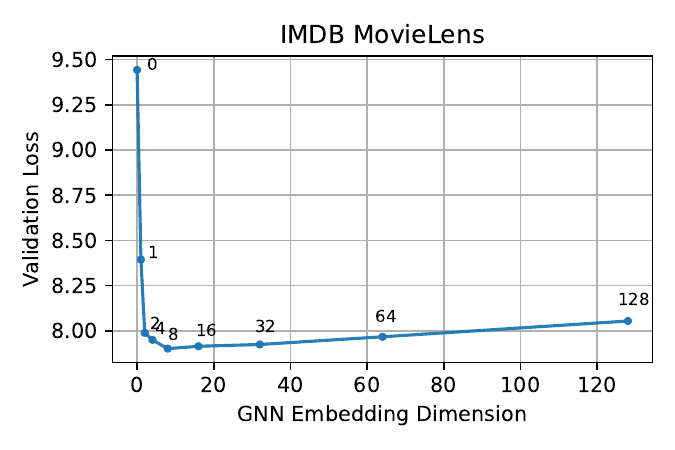}
        \caption{IMDB MovieLens}  
        \label{fig:sub1}
    \end{subfigure}
    \hfill
    \begin{subfigure}[b]{0.45\textwidth}
        \includegraphics[width=\textwidth]{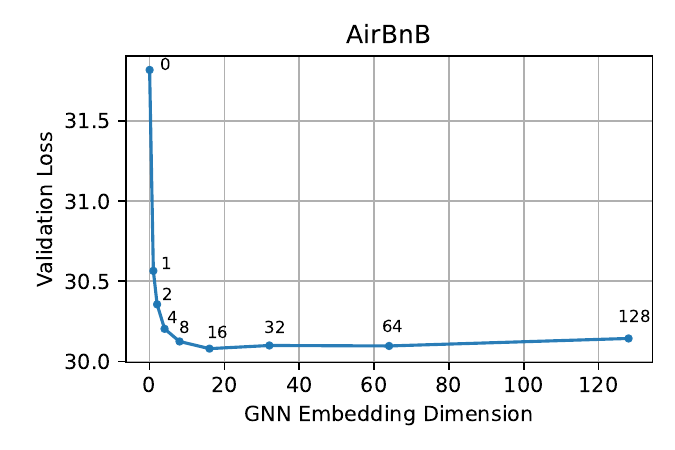}
        \caption{Airbnb}  
        \label{fig:sub2}
    \end{subfigure}
    \caption{Best validation loss as a function of the GNN embedding size. A value of zero means the GNN is not used.}
    \label{fig:emb_size_vs_val_loss}  
\end{figure}

The size of the embedding produced by the GNN is an important hyperparameter. A large embedding size can cause the neural network to memorize structures, while an overly small one will limit expressiveness by restricting information flow. For our experiments, we chose embedding sizes in the range of 2–10, a conservative choice to avoid overfitting and privacy leaks. Figure \ref{fig:emb_size_vs_val_loss} shows how the minimum validation loss varies with the embedding size for two datasets.

We observe that for the IMDB-MovieLens dataset, performance degrades if the embedding size is arbitrarily increased. For the Airbnb experiments, the effect is less pronounced, likely due to the different GNN architecture used.

\section{Datasets}
\label{app:datasets}

We summarize the datasets used in our experiments in Table~\ref{tab:datasets}. To allow comparisons with prior work, the AirBnB, Rossmann, and Walmart datasets were downsampled. AirBnB and Rossmann are relatively simple, each containing a single foreign key relationship, while the Walmart dataset features a table with two child tables. The IMDB, Biodegradability, and CORA datasets have more complex structures, including multiple foreign keys and tables referencing multiple parent tables. In particular, both Biodegradability and CORA include tables with two distinct foreign keys pointing to the same parent table. For example, the \texttt{cites} table in CORA contains both the ID of the citing paper and that of the cited paper.
Additional details about the datasets are provided in \citet{hudovernik2024benchmarking}.

\begin{table*}[ht]
\centering
\begin{tabular}{l l r r l}
\toprule
\textbf{Dataset} & \textbf{Table} & \textbf{\# Rows} & \textbf{\# Features} & \textbf{Foreign Keys} \\
\cmidrule(lr){1-5}
\multirow{2}{*}{AirBnB} 
  & users    & 10,000  & 15 & -- \\
  & sessions & 47,217  & 5  & users \\
\addlinespace
\cmidrule(lr){1-5}
\multirow{5}{*}{Biodegradability} 
  & molecule & 328     & 3  & -- \\
  & group    & 1,736   & 1  & -- \\
  & atom     & 6,568   & 1  & molecule \\
  & gmember  & 6,647   & --  & atom, group \\
  & bond     & 6,616   & 1  & atom1, atom2 \\
\addlinespace
\cmidrule(lr){1-5}
\multirow{3}{*}{CORA} 
  & paper    & 2,708   & 1  & -- \\
  & content  & 49,216  & 1  & paper \\
  & cites    & 5,429   & --  & paper1, paper2 \\
\addlinespace
\cmidrule(lr){1-5}
\multirow{7}{*}{IMDB MovieLens} 
  & users            & 6,039    & 3  & -- \\
  & movies           & 3,832    & 4  & -- \\
  & actors           & 98,690   & 2  & -- \\
  & directors        & 2,201    & 2  & -- \\
  & ratings          & 996,159  & 1  & movie, user \\
  & movies2actors    & 138,349  & 1  & movie, actor \\
  & movies2directors & 4,141    & 1  & movie, director \\
\addlinespace
\cmidrule(lr){1-5}
\multirow{2}{*}{Rossmann} 
  & store      & 1,115   & 9  & -- \\
  & historical & 57,970  & 7  & store \\
\addlinespace
\cmidrule(lr){1-5}
\multirow{3}{*}{Walmart} 
  & stores    & 45     & 2  & -- \\
  & features  & 225    & 11 & store \\
  & depts     & 15,047 & 4  & store \\
\bottomrule
\end{tabular}
\caption{\small Overview of the datasets used in the experiments, showing for each table of each dataset the number of rows, the number of features (feature columns) and tables referred by foreign keys (the parent tables).}
\label{tab:datasets}

\end{table*}

Finally, the sampling procedure of the foreign-key graph depended on the dataset. For the IMDB-MovieLens and the CORA datasets, where the largest connected component includes respectively $99\%$ and $93\%$ of the data, the original foreign-key graph was retained. For the other datasets, the graph was built by sampling with replacement an equal number of connected components from the original graph.

\end{document}